\documentclass[11pt]{article}

\usepackage{amscd,amssymb,stmaryrd}
\usepackage{graphicx}
\usepackage{subcaption}
\usepackage[utf8]{inputenc}
\usepackage[T1]{fontenc}    % use 8-bit T1 fonts
\usepackage[export]{adjustbox}
\usepackage[english]{babel}
\usepackage{cite}
\usepackage{color}
\usepackage{mathtools}
\usepackage{hyperref}
\usepackage[english]{babel}
\usepackage{booktabs}
\usepackage{float}

\usepackage{tikz}
\usepackage{pifont}% 

\usepackage{diagbox}
\usetikzlibrary{decorations.pathreplacing,calligraphy}
\usepackage{pgfplots}
\pgfplotsset{compat=1.11}
\usepgfplotslibrary{groupplots}

\usepackage{url}
\usepackage{graphicx}

\usepackage{hyperref}       % hyperlinks
\usepackage{url}            % simple URL typesetting
\usepackage{booktabs}       % professional-quality tables
\usepackage{amsfonts}       % blackboard math symbols
\usepackage{nicefrac}       % compact symbols for 1/2, etc.
\usepackage{microtype}      % microtypography
\usepackage{xcolor}         % colors

\usepackage{amsmath}
\usepackage{bm}
\usepackage{macros}
\usepackage{multirow}
\usepackage{nicematrix}
\title{PROSE-FD: A Multimodal PDE Foundation Model for Learning Multiple Operators for Forecasting Fluid Dynamics}

\author{
Yuxuan Liu\thanks{Department of Mathematics, UCLA, Los Angeles, CA 90095.}\qquad
Jingmin Sun\thanks{Department of Mathematical Sciences, Carnegie Mellon University, Pittsburgh, PA 15213.}\qquad
Xinjie He\footnotemark[1]\\
Griffin Pinney\footnotemark[1]\qquad 
Zecheng Zhang\thanks{Department of Mathematics, Florida State University, Tallahassee, FL 32304.}\qquad
Hayden Schaeffer\footnotemark[1]  }

\date{}

\begin{document}

\maketitle

\begin{abstract}
We propose PROSE-FD, a zero-shot multimodal PDE foundational model for simultaneous prediction of heterogeneous two-dimensional physical systems related to distinct fluid dynamics settings. These systems include shallow water equations and the Navier-Stokes equations with incompressible and compressible flow, regular and complex geometries, and different buoyancy settings.
This work presents a new transformer-based multi-operator learning approach that fuses symbolic information to perform operator-based data prediction, i.e. non-autoregressive. By incorporating multiple modalities in the inputs, the PDE foundation model builds in a pathway for including mathematical descriptions of the physical behavior.
We pre-train our foundation model on 6 parametric families of equations collected from 13 datasets, including over 60K trajectories. 
Our model outperforms popular operator learning, computer vision, and multi-physics models, in benchmark forward prediction tasks. 
We test our architecture choices with ablation studies. 
\end{abstract}

\let\thefootnote\relax\footnotetext{The code is available at: \url{https://github.com/felix-lyx/prose}.}

\section{Introduction}

Fluid dynamics and related physical models are essential for describing a wide range of scientific phenomena and are employed in various applications, including aerodynamics and aircraft design, weather forecasting, petroleum flow, space plasma dynamics and safety, combustion, and more. Simulating and predicting fluid dynamics is often complex and computationally expensive due to the system's highly nonlinear and multiscale behavior, such as the chaotic effects seen in turbulent flows. This challenge is amplified in real-world applications where measurements of state variables are scarce and noisy, some physical variables are unobserved, and boundary effects are unknown. Consequently, a key task in scientific machine learning is to develop models capable of learning general solution operators for fluid dynamics that can handle these issues.

Operator learning methods are a popular approach to train neural networks as surrogate models for solutions of partial differential equations (PDEs). These methods aim to train deep neural networks (DNNs) to approximate the map from input functions, such as boundary data and initial states, to the solution of the physical system. One advantage of neural operators is their potential for improved cost efficiency during inference \cite{pathak2022fourcastnet, sun2024lemon}. Recent advancements in operator learning include the Deep Operator Network (DeepONet) \cite{lu2019deeponet,lu2021learning, lin2023b} and the Fourier Neural Operator (FNO) \cite{li2020fourier}, which have demonstrated promise in scientific applications such as fluid dynamics \cite{li2022fourier} and weather prediction \cite{pathak2022fourcastnet}.

A key challenge with operator learning methods is that they are designed to train a single-operator network for one physical system at a time. Since training a deep neural network requires a large amount of data, this process often necessitates costly simulations or experiments. Moreover, because the operator network is trained on a single physical system, the resulting models do not exhibit emergent generalizations of physical properties \cite{sun2024towards}. PDE foundation models have emerged as a potential solution to address this by incorporating multiple physical systems into one model.

Foundation models in natural language processing and computer vision are deep learning models trained for multiple tasks using large datasets sampled from heterogeneous sources \cite{bommasani2021opportunities}. Approaches such as BERT \cite{devlin2018bert}, GPT \cite{radford2018improving, radford2019language, brown2020language}, DALL-E \cite{ramesh2021zero,ramesh2022hierarchical}, Stable Diffusion \cite{rombach2022high}, and LLAMA \cite{touvron2023llama, touvron2023llama2}, as well as Claude, have demonstrated success in data processing and generative tasks, showing evidence of generalization to new downstream tasks. However, these models have not been directly applicable to scientific computing problems, such as solving forward and inverse problems in PDEs, which require a higher degree of accuracy.

PDE foundation models aim to approximate solution operators for large classes of PDEs within a single DNN. The goal is to learn a general operator that can represent and infer the forward dynamics of distinct physical systems within one model. Thus, the fundamental task is to develop and train a DNN to accurately generalize to unseen physical dynamics that may share features with the trained dynamics. Current PDE foundation models include Predicting Operators and Symbolic Expressions (PROSE) \cite{liu2023prose,sun2024towards, sun2024lemon}, In-Context Operator Network (ICON) \cite{yang2023context, yang2023prompting, yang2024pde}, Multiple Physics Pretraining (MPP) \cite{mccabe2023multiple}, Fourier Forecasting Network (FourCastNet) \cite{pathak2022fourcastnet}, and Aurora \cite{bodnar2024aurora}.

PROSE is a multimodal PDE foundation model that simultaneously learns to predict the values of state variables and derives a symbolic formulation of the governing equations describing the physical system \cite{schaeffer2017learning, schaeffer2017sparse, sun2020neupde}. PROSE has been applied to ordinary differential equations with chaotic behavior \cite{liu2023prose} and one-dimensional time-dependent nonlinear PDEs \cite{sun2024towards}, and can incorporate robust fine-tuning and meta-learning strategies \cite{sun2024lemon}. In \cite{sun2024towards,sun2024lemon}, it was shown that PROSE can generalize physical features to unseen conservation laws, though its capabilities in higher-dimensional PDE systems remain open. ICON employs in-context learning to guide the model in predicting state variables based on examples, which has been shown to generalize predictions for one-dimensional conservation laws. 
MPP pre-trains an autoregressive vision transformer \cite{dosovitskiy2020vit, efendiev2022efficient} to map observations to future states, although the formulation can be unstable for long prediction windows. For weather prediction, FourCastNet uses the Adaptive Fourier Neural Operator model \cite{guibas2021adaptive}, while Aurora utilizes the 3D Swin Transformer \cite{liu2021swin} to generate higher-resolution predictions. However, both models specialize in atmospheric forecasting, may lose some high-frequency information, and are tailored to specific problems.

\textbf{Main Contributions:} 
We present \textit{PROSE-FD}, a pre-trained PDE foundational model that uses a new transformer-based deep neural network that leverages state-variable observations and symbolic information to perform operator-based data prediction for fluid dynamics (FD). The model's formulation allows for the inclusion of various sources of information, including mathematical equations that describe the governing physics, in the inputs and/or outputs. Our contributions are listed below.
\begin{itemize}
    \item We develop a multimodal fluids foundation model for predicting solution operators for shallow water equations and the Navier-Stokes equations with incompressible and compressible flow, regular and complex geometries, and different buoyancy settings.
    \item We show that the PROSE-FD model is capable of accurate predictions for fluid dynamics with a range of physical behavior.
\item Using 13 datasets, we demonstrate that PROSE-FD is able to outperform single-operator learning approaches, computer vision models, and other multi-physics models, in the prediction and forecasting tasks with gains ranging from 1.3x to 7.9x.
    \item Our code and parameters are open-sourced for future experimentation and comparisons. 
\end{itemize}

\section{Methods}

The main components of PROSE-FD include patch-based data encoding and decoding, symbolic equation encoding, and multimodal information fusion. We provide the problem description and the key components of PROSE-FD. We begin by summarizing related multimodal works.

\subsection{Multimodal Machine Learning (MMML)}

One key aspect of foundation models is that they are capable of multimodal machine learning (MMML) \cite{lu2019vilbert,sun2019videobert, tan2019lxmert,li2021ai,xu2023multimodal,liang2022foundations}, which focuses on building neural networks with the ability to comprehend, reason, and learn from diverse data sources and structures. As an example, for reasoning in image captioning generation, a multimodal model learns both visual features from the image and additional information from the corresponding textual data \cite{pmlr-v37-xuc15,nam2017dual}. While the success of foundation models in text-based tasks is well-established, their ability to reason and accurately represent quantities in scientific computing (SC) is limited. This is primarily due to the nature of the data, i.e. scientific data has lower information density than text, and due to the high precision requirements in SC problems. Specifically, there is a demand for a level of accuracy in scientific predictions that general-purpose foundation models are currently not designed to handle effectively.  Some key challenges in MMML include (1) representation learning and (2) reasoning and generation.
For representation learning, distinct but cooperative information from different modalities must be integrated into a uniform representation. PROSE-FD addresses this by utilizing a fusion layer with self-attention \cite{vaswani2017attention,bahdanau2014neural,xu2015show} sub-layers to enable information exchange between the two modalities for a holistic representation. For reasoning and generation, a model must provide a comprehensive understanding of the information gathered from different stages, such as representations from the fusion process and query locations. Cross-attention layers in the decoders facilitate this exchange and strengthen inter-modality relationships.

\subsection{Problem Setting}
Consider parametric families of two dimensional time-dependent nonlinear PDEs, whose state-variables of interest are represented by $\u(\x,t) \in \mathbb{R}^d$ (for some $d$ up to 4 in our tests) with $\x\in \Omega\subseteq \R^2$. 
Given data up to time $T_0$, i.e. the sequence: \[\{\u(\cdot,t_i)\st 0\le i < T_0\},\] the goal of the forward problem is to predict the subsequent $T$ timestamps \[\{\u(\cdot,t_i)\st T_0\le i<T_0+T\}.\] More generally, for operator learning, the goal is to learn the solution as a map $(\x, t)\mapsto \u(\x,t)$ for $(\x,t)\in \Omega \times [T_0 \Delta t, (T_0 + T) \Delta t]$ where $\Delta t = t_{i+1} - t_i$ is the timestep. In our experiments, we set $T_0 = T = 10$, that is, 10 timestamps are given as inputs and we predict, as operator evaluations, 10 future timestamps.

\subsection{Model Overview}
\begin{figure}
    \centering
    \includegraphics[width=\linewidth]{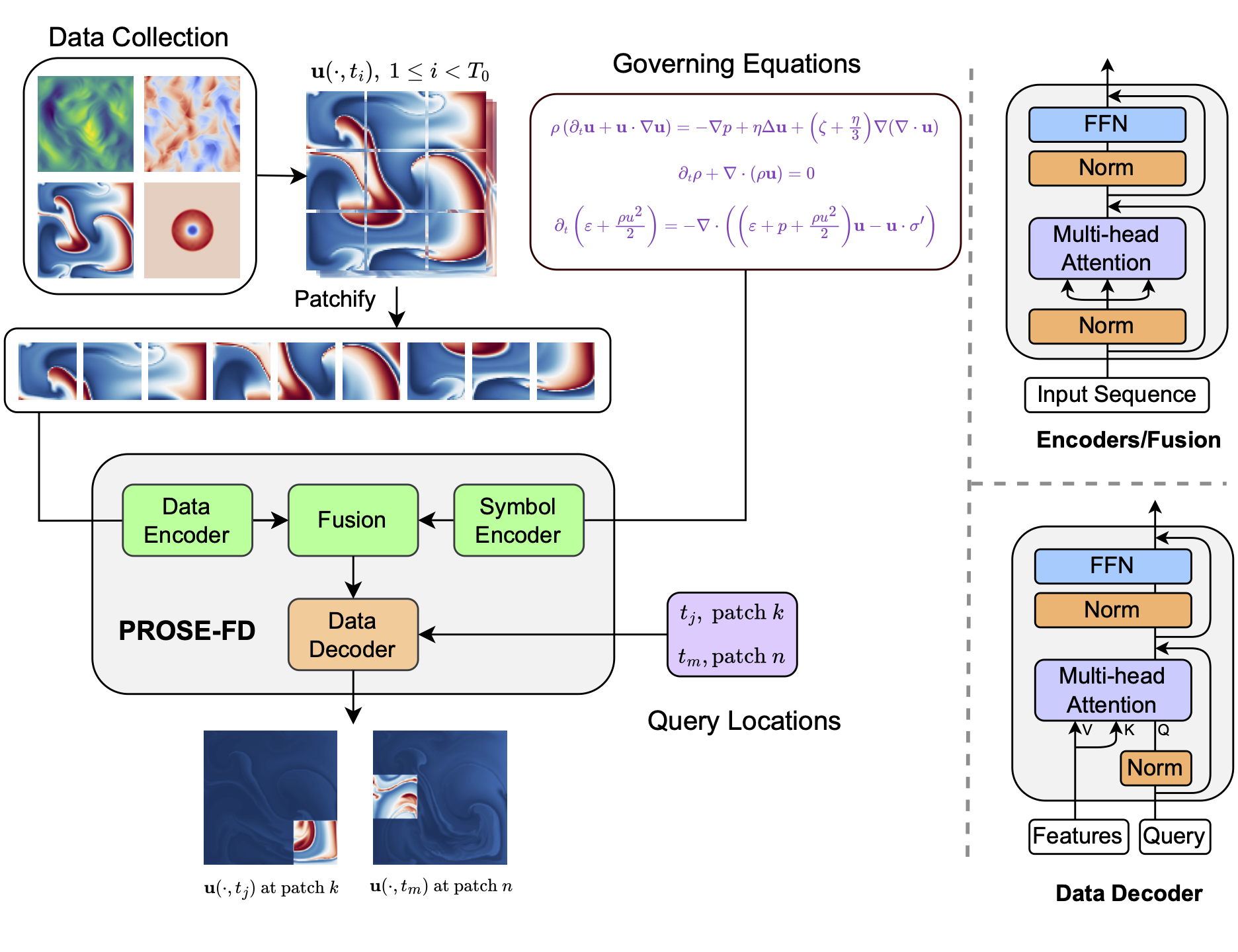}
    \caption{\textbf{PROSE-FD model overview.} The inputs to the model are the trajectories and governing equations (in symbolic form) sampled from the datasets. The input data is patchified before being converted into a sequence of features, which are then processed with encoders and fused with processed symbolic information. The data decoder takes in fused features and generates predictions at given query locations.}
    \label{fig:main_model}
\end{figure}

We use transformer layers as the backbone of the PROSE-FD model. In Figure \ref{fig:main_model}, we provide an illustration of our model. The input data $\{\u(\cdot,t_i)\st 0\le i < T_0\}$ is first converted into patches before being mapped into a sequence of tokens, which is then processed with the Data Encoder consisting of transformer layers. The input equation symbols are tokenized into a sequence of word embeddings, which are then processed by the Symbol Encoder and fused with data input in Fusion. The Data Decoder takes queries (based on $t$ and the patch) along with the fused features to generate the solution at the specified locations. Notably, the model's computational complexity is linear with respect to the number of query locations, as each query is evaluated independently.

\subsection{Patch-based Data Encoder}
Processing the input data using transformers requires first converting the input into a sequence of tokens. For ODE systems or 1D PDE, directly projecting each timestamp into a token is an effective approach \cite{liu2023prose, sun2024towards}. However, for 2D PDE, a similar approach will lead to information loss due to the curse-of-dimensionality: for space resolution $128\times 128$ and 3 channels (e.g. velocities and pressure), each token needs to encode information of dimension $128^2\cdot 3\approx 50K$, which is much larger than the hidden dimension. 
Patch-based encoding strategies were used in ViT \cite{dosovitskiy2020vit} to balance the sequence length and token dimension. More precisely, each image (i.e. each timestamp $\u(\cdot,t)$) is first converted into $p^2$ patches each of resolution $(128/p)\times (128/p)$. These patches are then transformed and flattened into a sequence of $p^2$ tokens. We use $p=8$, thus each token encodes spatial information of dimension $16^2\cdot 3 =768$, which is smaller than 1024, the hidden dimension of the linear and attention layers. Consequently, spatially dependent information is retained at the cost of a longer sequence. 

We choose $p = 8$ for our inputs, thus the data input is first converted into a sequence of $64\cdot T_0$ tokens. After adding learnable timestamps and patch positional encodings, the sequence is further processed using self-attentions. 
Compared to other video transformer models such as Axial attention \cite{ho2019axial}, this approach makes it easier to fuse information from different modalities. 

\subsection{Equation Encoding and Fusion}
An important aspect of the simultaneous learning of multi-operator solution operators is to encode the equations so that commonalities and differences can be automatically detected and processed. We encode the equations as symbolic trees \cite{liu2023prose, sun2024towards} (operations and functions as nodes, variables and constants as leaves), which are then converted to a sequence in Polish notation. The sequence is then processed similarly to sentences: they are converted to trainable tokens and then further processed using self-attention. The tokens retain meaning, i.e. mathematical functions such as ``tan'' and ``add'' are not further tokenized. 
Compared to \LaTeX ~encoding of the equation, this approach has a shorter sequence length and easier syntax. For more details, we refer to \cite{jiang2023finite,d2022deep,kamienny2022end}. 

To fuse information obtained from two input modalities, the processed data and symbol sequence are concatenated into a single sequence and further processed through self-attention in the Feature Fusion block. By attending to the symbols, the data sequence obtains information from the symbolic input (e.g., aspects of the equation underlying the data).

\subsection{Patch-based Operator Decoder}
For operator learning, the usual approach is to construct the solution as the map $(\x,t)\mapsto \u(\x,t)$.
The PROSE-FD model constructs the solution via cross-attention, where the points $(\x,t)$ serve as queries, and the encoded features serve as keys and values. As a result of the curse-of-dimensionality issue, constructing the solution for all $(\x,t)$ with cross-attention is computationally expensive due to the increase in sequence length, i.e. the number of spatial points became the number of elements in the sequence. To ensure the independence of the query evaluation while maintaining reasonable computational complexity, instead of constructing a function evaluated for each input spacial location $\x$, the Data Decoder block learns a function that maps patches $P$. That is, given a patch $P$ representing a set of spatial coordinates $P = \{\x_k\}_{k\in K}$ (where $K$ is an index set with cardinality of 64 in our applications), the Data Decoder learn the solution $(P,t)\mapsto \{\u(\x_k,t_i)\st \x_k\in P\}.$ 
Additionally, due to the linear complexity of the Data Decoder, the output sequence length can be larger than the encoder's sequence length. Consequently, we use $p=16$ output patches in each dimension.

\section{Experiments}
In this section, we first explain the experiment setup. We then present the main results and compare our PROSE-FD model with other baseline models. Finally, we validate our key architecture choices with ablation studies. More experiment details can be found in Appendix \ref{sec:exp_details}.

\subsection{Experiment Setup}

\paragraph{Dataset.} The dataset we use contains 6 parametric families of PDEs modeling fluid dynamics in different regimes collected from 3 heterogeneous sources: PDEBench \cite{takamoto2022pdebench}, PDEArena \cite{gupta2022towards}, and CFDBench \cite{luo2023cfdbench}. The dataset includes shallow water equations and the Navier-Stokes system with incompressible and compressible flow, regular and complex geometries, and different buoyancy settings. For datasets that do not provide a train/val/test splitting, we use the standard 80\%/10\%/10\% splitting.  For more details, we refer to Appendix \ref{sec:dataset_details}.

\paragraph{Evaluation Metric.}
The relative $L^2$ norm is used as the evaluation metric. More precisely, given the model's prediction $\tilde{\u}$ and the ground truth $\u$, we compute the (time-averaged) relative $L^2$ error: 
\begin{equation}
    \frac{1}{T}\sum_{i=T_0}^{T_0 + T-1}\frac{\|\u(\cdot,t_i) - \tilde{\u}(\cdot,t_i)\|_2}{\|\u(\cdot,t_i)\|_2 + \ep}  ,
\end{equation} 
where $T_0=10$ is the number of input steps, $T=10$ is the number of output steps, and $\ep = 10^{-7}$. 
Note that for the Navier-Stokes dataset from PDEArena, the temporal grid resolution is only 14, thus we set $T=4$ for this dataset only. The average used in the last column of Table~\ref{tab:main_results} is the average of the relative $L^2$ errors over the 6 families, i.e. the average over each row of the table. 

\subsection{Baselines and Comparisons}
We compare our PROSE-FD model with the following baselines. DeepONet \cite{lu2019deeponet} and FNO \cite{li2020fourier} are popular single-operator learning methods that efficiently approximate PDE solution operators. UNet \cite{ronneberger2015u} is a classical convolution-based image processing model, utilizing symmetric hierarchical structures to capture both context and fine details for pixel-wise predictions. ViT \cite{dosovitskiy2020vit} is a popular transformer-based image processing model that captures global image dependencies and demonstrates scalability for model sizes in image and video processing tasks. MPP \cite{mccabe2023multiple} is an Axial-ViT-based multi-physics pretraining approach, which autoregressively predicts PDE solutions. More details about the baselines are included in Appendix \ref{sec:baseline_details}.

\subsection{Main Results}
\begin{figure}
\centering
\begin{subfigure}[b]{\textwidth}
   \includegraphics[width=1\linewidth]{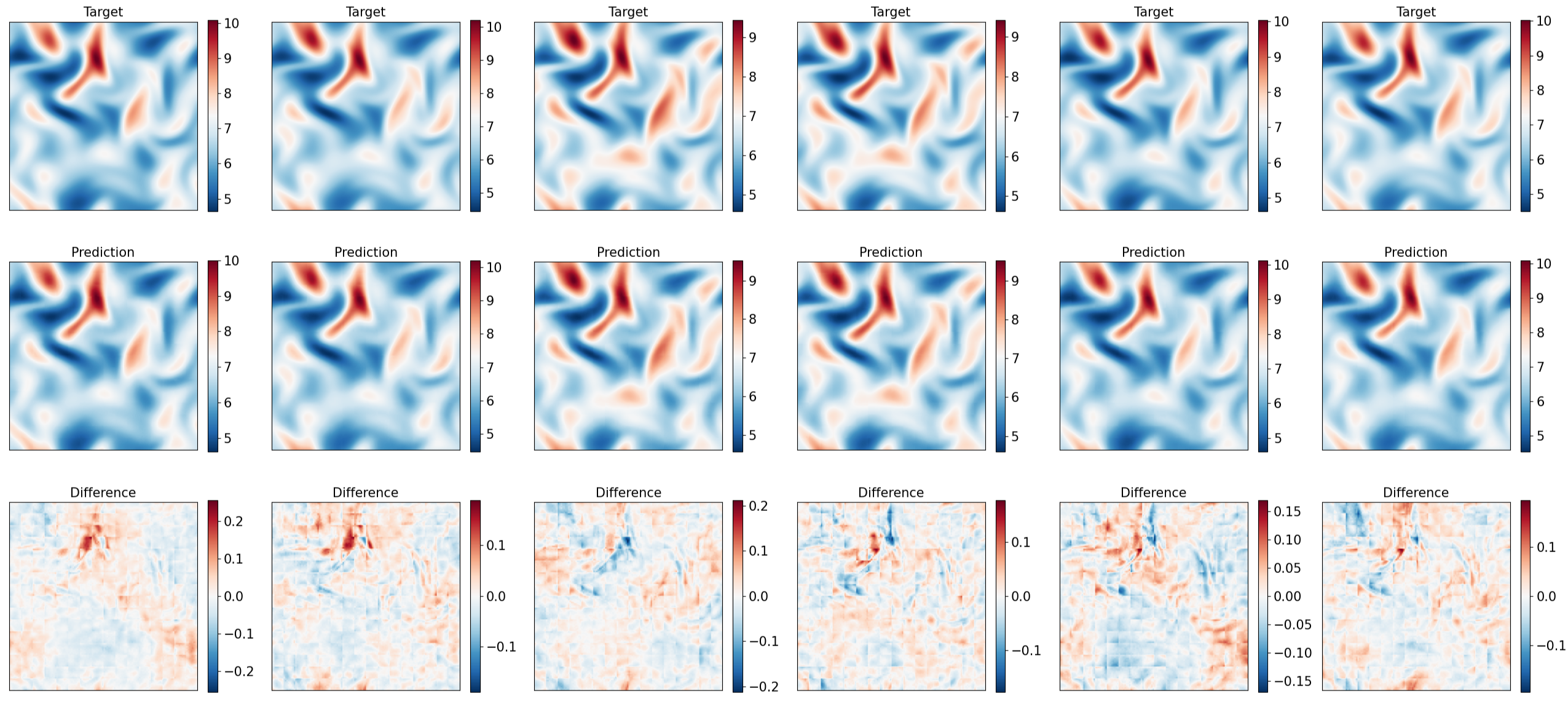}
   \caption{5 output steps for PDEBench Compressible Navier-Stokes dataset. The channel plotted is the density field in equation \eqref{eq:cns}. Each column represents a different timestamp. For this trajectory, the relative $L^2$ error is 0.34\%.\\}
\end{subfigure}
\begin{subfigure}[b]{\textwidth}
   \includegraphics[width=1\linewidth]{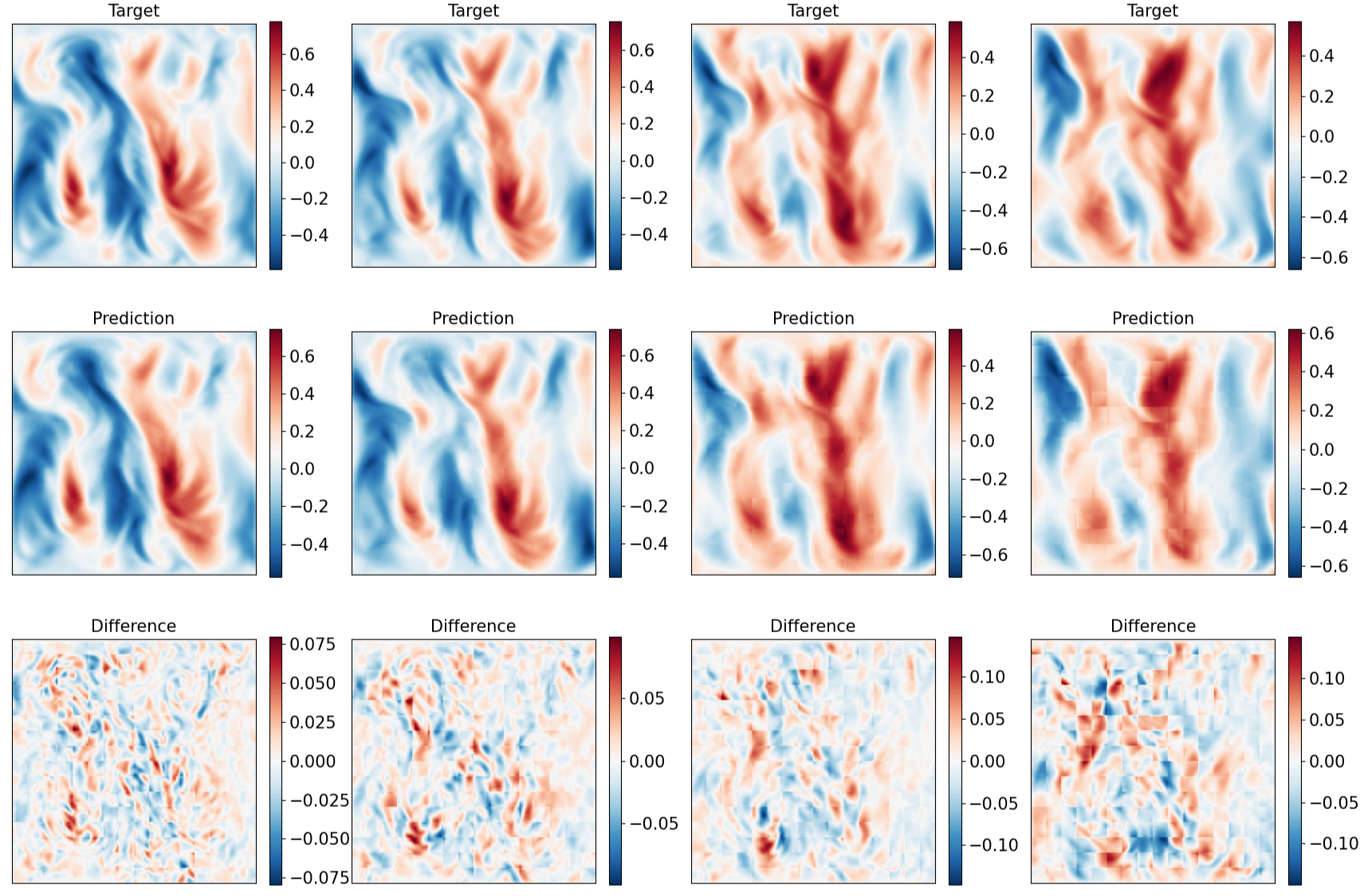}
   \caption{4 output steps for PDEArena Navier-Stokes dataset. The channel plotted is the x-velocity in equation \eqref{eq:arena_ns}, i.e. $u_x$. Each column represents a different timestamp. For this trajectory, the relative $L^2$ error is 5.13\%. This examples shows that even when the relative error is higher, the structures of the flow can still be predicted correctly.}
\end{subfigure}

\caption{\textbf{Two example outputs for the PROSE-FD model.}}
\label{fig:ex_output}
\end{figure}

The main experiment results are included in Table \ref{tab:main_results}, where we report the relative $L^2$ error (\%) for each family of equations and the average. For all the models, we use the same training setting: a single model is trained to predict all families of equations, without any fine-tuning. Our PROSE-FD model exhibits remarkable performance, outperforming all baselines in all but one family of equations. 
We include example visualizations of the PROSE-FD model output in Figure \ref{fig:ex_output} and Appendix \ref{sec:more_visual}.

\begin{table}[]
\centering
\caption{\textbf{Main Results and Comparisons with Baselines}. The numbers reported are relative $L^2$ errors (\%). The averages are taken with respect to the 6 distinct families listed in the columns of the table. For each family of equations, we \textbf{bold} the best results. *Note that the PDEBench CNS contains 8 subsets of parameter configurations.}
\label{tab:main_results}
{
\footnotesize
\begin{NiceTabular}{cc|cccccc|c}
\toprule
\multirow{2}{*}{Model} & \multirow{2}{*}{Param} & \multicolumn{3}{c|}{PDEBench}        & \multicolumn{2}{c|}{PDEArena}     & CFDBench & \multirow{2}{*}{Average} \\
                       &                        & SWE & CNS* & \multicolumn{1}{c|}{INS} & NS & \multicolumn{1}{c|}{NS-cond} & -      &                          \\ 
\midrule
FNO      & 0.6M & 3.71 & 6.31 & 36.83 & 38.68 & 55.63 & 8.53 & 24.95 \\
DeepONet & 3.5M & 3.55 & 7.41 & 64.61 & 35.33 & 51.85 & 12.50 & 29.21 \\
UNet     & 5.6M & 0.33 & 3.19 & 3.43 & 12.56  & 16.82 & 0.76 & 6.18 \\
ViT      & 154M & 0.30 & 2.70 & 3.14 & 10.19 & 15.71 & 0.70 & 5.34 \\
MPP-B    & 116M & 1.02 & 1.90 & 7.52 & \textbf{5.71} & 12.56 & 1.23 & 4.99 \\
PROSE-FD    & 169M & \textbf{0.28} & \textbf{1.53} & \textbf{2.84} & 6.34 & \textbf{10.76} & \textbf{0.54} & \textbf{3.71} \\ 
\bottomrule
\end{NiceTabular}
}
\end{table}

\subsection{Ablation Studies}

In this section, we present the results of our ablation studies to validate some key architecture choices. We compare the zero-shot testing performance (relative $L^2$ error) averaged over the 6 datasets. The results are shown in Table \ref{tab:ablation}.

\paragraph{Rollout vs. Operator Formulation.}
Our baseline PROSE-FD model learns the map from $T_0$ input timestamps to $T$ output steps in a non-autoregressive way, i.e., it can output multiple future timestamps in a single forward pass. The Data Decoder is an operator in time and no explicit rollout is needed. An alternative strategy is to have the model learn the map from $T_0$ input timestamps to 1 output step, then rollout, i.e. recursively apply, the model during the inference stage \cite{efendiev2022efficient}. To obtain $T$ output steps, the model needs to be called $T$ times to generate the full solution. As shown in Table \ref{tab:ablation}, our baseline model outperforms the rollout model, demonstrating that the operator learning approaches can avoid error accumulation.

\paragraph{Influence of Symbolic Information.}
To show the importance of symbolic equation information, we compare our baseline PROSE-FD model with a model that only uses the Data Encoder and the Data Decoder. We increase the number of layers for each component so the final parameter count is close to the baseline model. 
The comparison in Table \ref{tab:ablation} demonstrates that the symbolic encoding structure of the PROSE-FD model enhances the predictions.
\begin{table}[]
    \caption{\textbf{Results for PROSE-FD Ablation Studies.} We compare the baseline (operator) approach to a rollout (recurrent) prediction variant and a data-only variant.}
    \label{tab:ablation}
    \centering
    \begin{tabular}{c c c}
    \toprule
    Model & Param & Average Testing Relative $L^2$ Error (\%) \\
    \midrule
    Baseline & 169M & 3.71 \\
    Rollout in time & 169M & 3.74 \\
    Data-only, no symbolic information & 168M & 4.06 \\
    \bottomrule
    \end{tabular}
\end{table}

\section{Discussion}
PROSE-FD is a pre-trained PDE foundational model that utilizes transformers to encode and process data and symbolic information for predicting solutions for fluid systems. The multimodality aspect of the approach allows for further experimentation and enhancement by including additional modality information that describes the physical systems of interest.
Our model is able to encode information from the two-dimensional shallow water equations and the two-dimensional Navier-Stokes equations with incompressible and compressible flow, regular and complex geometries, and different buoyancy settings into one model. Through extensive testing, we demonstrated that the model's predictions accurately capture the behavior presented in the datasets used -- outperforming other single-operator learning and transformer models. Thus, the approach presents a general-purpose surrogate model for two-dimensional fluid systems. Further work will examine the scalability of the model and the encoding of boundary effects.

\subsubsection*{Acknowledgments}
Y. Liu, X. He, G. Pinney, and H. Schaeffer were supported in part by NSF 2331033 and AFOSR MURI FA9550-21-1-0084. Z. Zhang was supported by DE-SC0025440.

\newpage
\bibliography{references}
\bibliographystyle{plain}

\newpage
\appendix
\section{Dataset Details}\label{sec:dataset_details}
The data was obtained from the PDEBench \cite{takamoto2022pdebench}, PDEArena \cite{gupta2022towards}, and CFDBench \cite{luo2023cfdbench} datasets. Unless otherwise specified, the space resolution is $128\times 128$. 

\subsection{PDEBench \cite{takamoto2022pdebench}}
\paragraph{Shallow Water Equation.}
The quantity of interest is the water depth $h(\x,t)$ on domain $[-2.5,2.5]^2\times [0,1]$ with Neumann boundary condition. The temporal resolution is 101. The equations are: \begin{align}
    \p_t h + \nabla h\u &= 0,\\
    \p_t h\u + \nabla \left(h \,\u \cdot \u + \frac12 g_r h^2\right) &= - g_r h \nabla b.
\end{align}

\paragraph{Incompressible Navier-Stokes Equation.}
The quantities of interest are the velocities $\u(\x,t)$ and particle density $c(\x,t)$ on domain $[0,1]^2\times [0,5]$ with Dirichlet boundary condition. The temporal resolution is 1000. The equations are: 
\begin{align}
    \rho(\p_t \u + \u \cdot \nabla \u) &= -\nabla p + \mu \Delta \u + \mathbf{F},\\
    \nabla \cdot \u &= 0,\\
    \p_t c + \nabla \cdot (c\u) &= 0.
\end{align} 
The forcing term $\mathbf{F}$ is randomly sampled.  

\paragraph{Compressible Navier-Stokes Equation.}
The quantities of interest are the velocities $\u(\x,t)$, pressure $p(\x,t)$, and density $\rho(\x,t)$ on domain $[0,1]^2\times [0,1]$ with periodic boundary conditions. The temporal resolution is 21. For equations with low viscosities, the dataset is provided on a finer $512\times 512$ space grid, which is downsampled to $128\times 128$ for consistency (through average pooling). The equations are:
\begin{align}\label{eq:cns}
    \partial_t \rho + \nabla \cdot (\rho \u) &= 0,\\
    \rho(\partial_t \u + \u\cdot \nabla \u) &= - \nabla p + \eta \Delta \u + (\zeta + \eta/3) \nabla (\nabla\cdot \u),\\
    \partial_t \left(\ep + \frac{\rho u^2}{2}\right) &= - \nabla \cdot \left(\left(\varepsilon + p + \frac{\rho u^2}{2}\right)\u - \u\cdot \sigma'\right).
\end{align}

\subsection{PDEArena \cite{gupta2022towards}}
\paragraph{Incompressible Navier-Stokes Equation.}
The quantities of interest are the velocities $\u(\x,t)$ and particle density $c(\x,t)$ on domain $[0,32]^2\times [18,102]$ with Dirichlet boundary conditions for velocity and Neumann boundary condition for particle field. The temporal resolution is 14. The equations are: \begin{align}\label{eq:arena_ns}
    \rho(\p_t \u + \u \cdot \nabla \u) &= -\nabla p + \mu \Delta \u + \mathbf{F},\\
    \nabla \cdot \u &= 0,\\
    \p_t c + \nabla \cdot (c\u) &= 0.
\end{align} The forcing term $\mathbf{F}$ takes the form $\mathbf{F} = (0, f)$ with $f=0.5$. 

\paragraph{Incompressible Navier-Stokes Equation (Conditioned).}
The quantities of interest are the velocities $\u(\x,t)$ and particle density $c(\x,t)$ on domain $[0,32]^2\times [18,102]$ with Dirichlet boundary conditions for velocity and Neumann boundary condition for particle field. The temporal resolution is 56. The equations are: \begin{align}\label{eq:arena_ns_c}
    \rho(\p_t \u + \u \cdot \nabla \u) &= -\nabla p + \mu \Delta \u + \mathbf{F},\\
    \nabla \cdot \u &= 0,\\
    \p_t c + \nabla \cdot (c\u) &= 0.
\end{align} The forcing term $\mathbf{F}$ takes the form $\mathbf{F} = (0, f)$ where $f$ is uniformly sampled in $[0.2,0.5]$. 

\subsection{CFDBench \cite{luo2023cfdbench}}
\paragraph{Incompressible Navier-Stokes Equation.}
The quantities of interest are the velocities $\u(\x,t)$ and pressure $p(\x,t)$. This dataset contains irregular geometries with Dirichlet boundary conditions. The raw space resolution is $64\times 64$ which is upsampled to $128\times 128$ via interpolation. The equations are: \begin{align}
    \rho(\p_t \u + \u \cdot \nabla \u) &= -\nabla p + \mu \Delta \u,\\
    \nabla \cdot \u &= 0.
\end{align}

\section{Experiment Details}\label{sec:exp_details}
We provide more details about the training process, architecture, and baselines. 

\subsection{Training}
We perform data normalization during the training process. Given the input sequence of data $\{\u(\cdot,t_i)\st 0\le i < T_0\}$, we compute the mean and standard deviation of each input trajectory, which are used to normalize both the input and ground truth sequence. The loss function is the standard mean squared error in the normalized space. The models are trained using the AdamW optimizer with a global batch size of 176 for 40 epochs where each epoch is 4,000 steps. The warmup-stable-decay learning rate scheduler \cite{hu2024minicpm} is used with 10\% warmup and 20\% decay. We use learning rate $10^{-4}$ and weight decay $10^{-4}$. On two NVIDIA H100 GPUs, the training takes about 58 hours. 

\subsection{Model Hyperparameters}
The model hyperparameters are summarized in Table~\ref{tab:model_hyper}.

\begin{table}[b]
    \centering
    \footnotesize
    \caption{\textbf{Model hyperparameters.} FFN means feedforward network.}
    \label{tab:model_hyper}
    \begin{tabular}{l l | l l }
    \toprule
    Hidden dimension - attention & 1024 & Hidden dimension - FFN & 2048\\
    Number of attention heads & 8 & Fusion attention layers & 8\\
    Data encoder attention layers & 2 & Data decoder attention layers & 8\\
    Symbol encoder attention layers & 4 & Dropout & 0 \\
    Input patch number & 8 & Output patch number & 16 \\
    \bottomrule
    \end{tabular}
\end{table}

\subsection{Baselines} \label{sec:baseline_details}
In this section, we include more details about the compared models.

\paragraph{DeepONet \cite{lu2019deeponet}.}
We employ the unstacked DeepONet architecture, consisting of a single trunk network and a single branch network. Initially, the input data is divided into $8 \times 8$ patches, with each patch being embedded into a 128-dimensional vector. These vectors are then passed through the branch network, producing an output with a basis dimension of $p = 50$. Simultaneously, the query point is processed through the trunk network, which also outputs a vector with the same dimension, $p$. The output solution at the query point is obtained by taking the inner product of the outputs from the two networks.

\paragraph{FNO \cite{li2020fourier}.}
We use 4 layers of standard 3d FNO to process the input data. The number of modes to keep in each dimension is set to 8, and the number of hidden channels is set to 16. 

\paragraph{UNet \cite{ronneberger2015u}.}
We use 8 layers of 3d UNet with GeLU activation and 32 hidden dimensions. 

\paragraph{ViT \cite{dosovitskiy2020vit}.}
For ViT, we use 10 layers of transformer encoder. The input patch number is set to be 8, the hidden dimension for attention is 1024, the hidden dimension for the feedforward network is 2048, and the number of heads is 8. 

\subsection{More Visualizations} \label{sec:more_visual}
\begin{figure}
\centering
\begin{subfigure}[b]{0.8\textwidth}
   \includegraphics[width=1\linewidth]{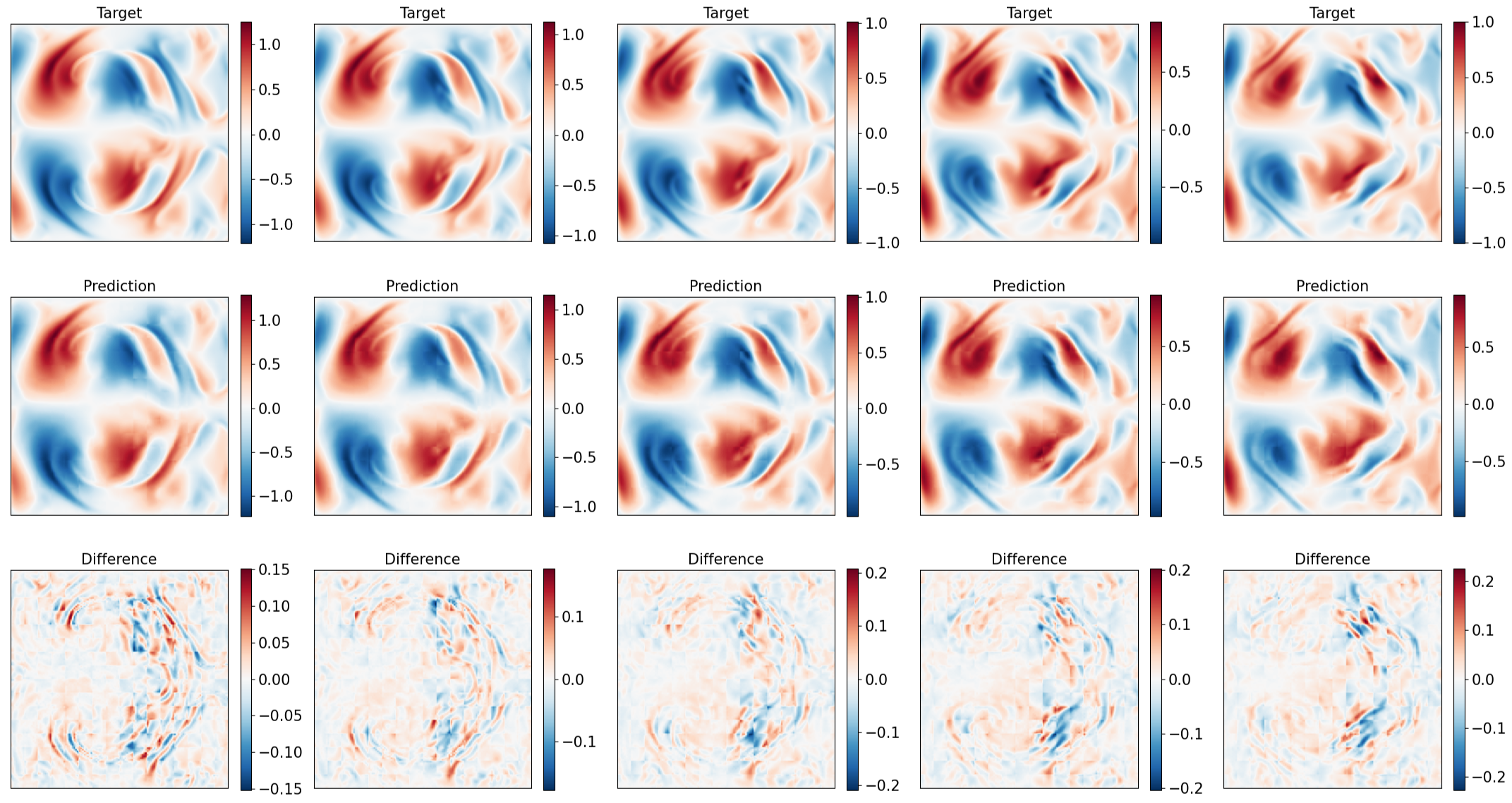}
   \caption{The channel plotted is the x-velocity.}
\end{subfigure}
\begin{subfigure}[b]{0.8\textwidth}
   \includegraphics[width=1\linewidth]{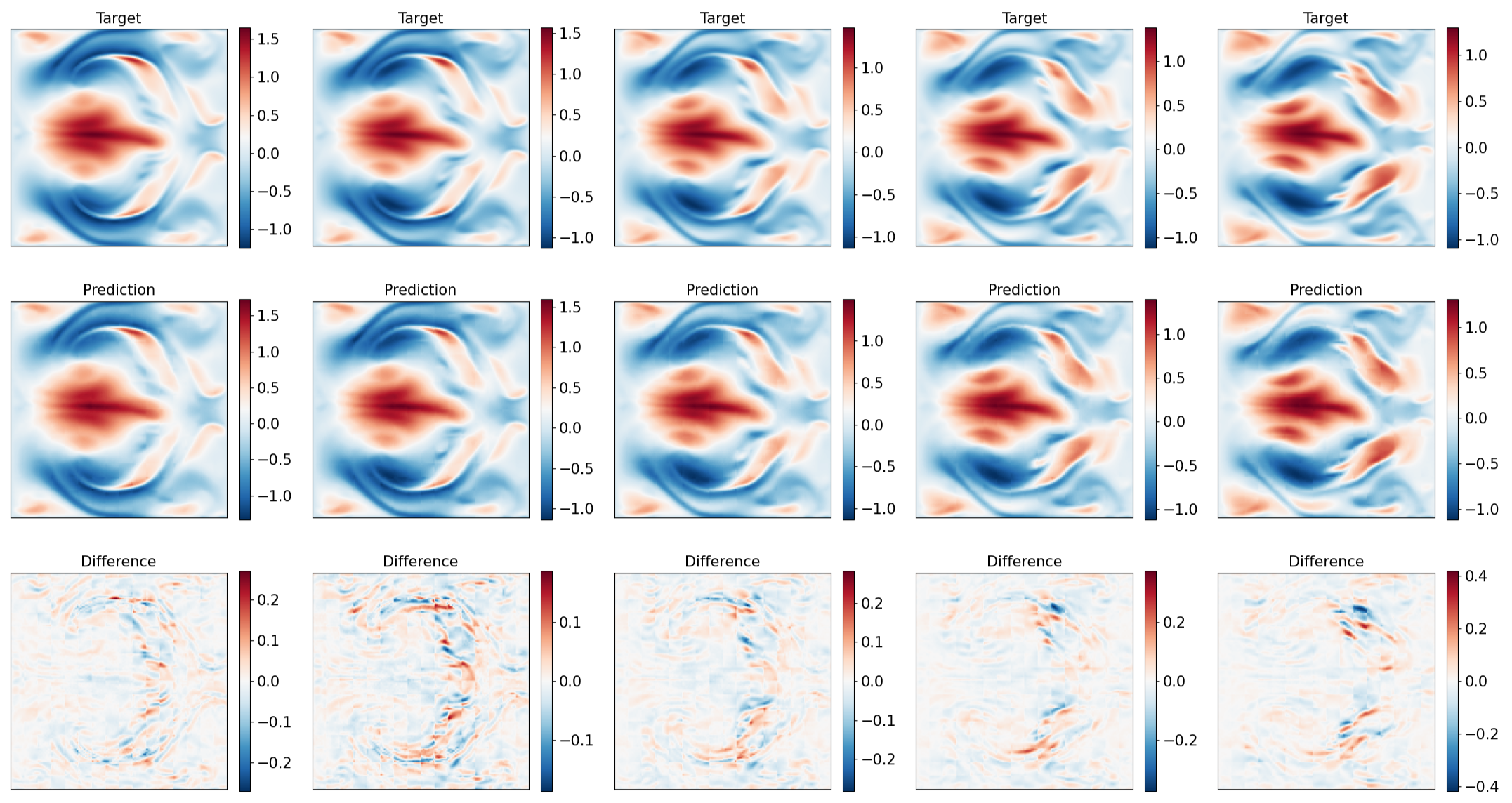}
   \caption{The channel plotted is the y-velocity.}
\end{subfigure}
\begin{subfigure}[b]{0.8\textwidth}
   \includegraphics[width=1\linewidth]{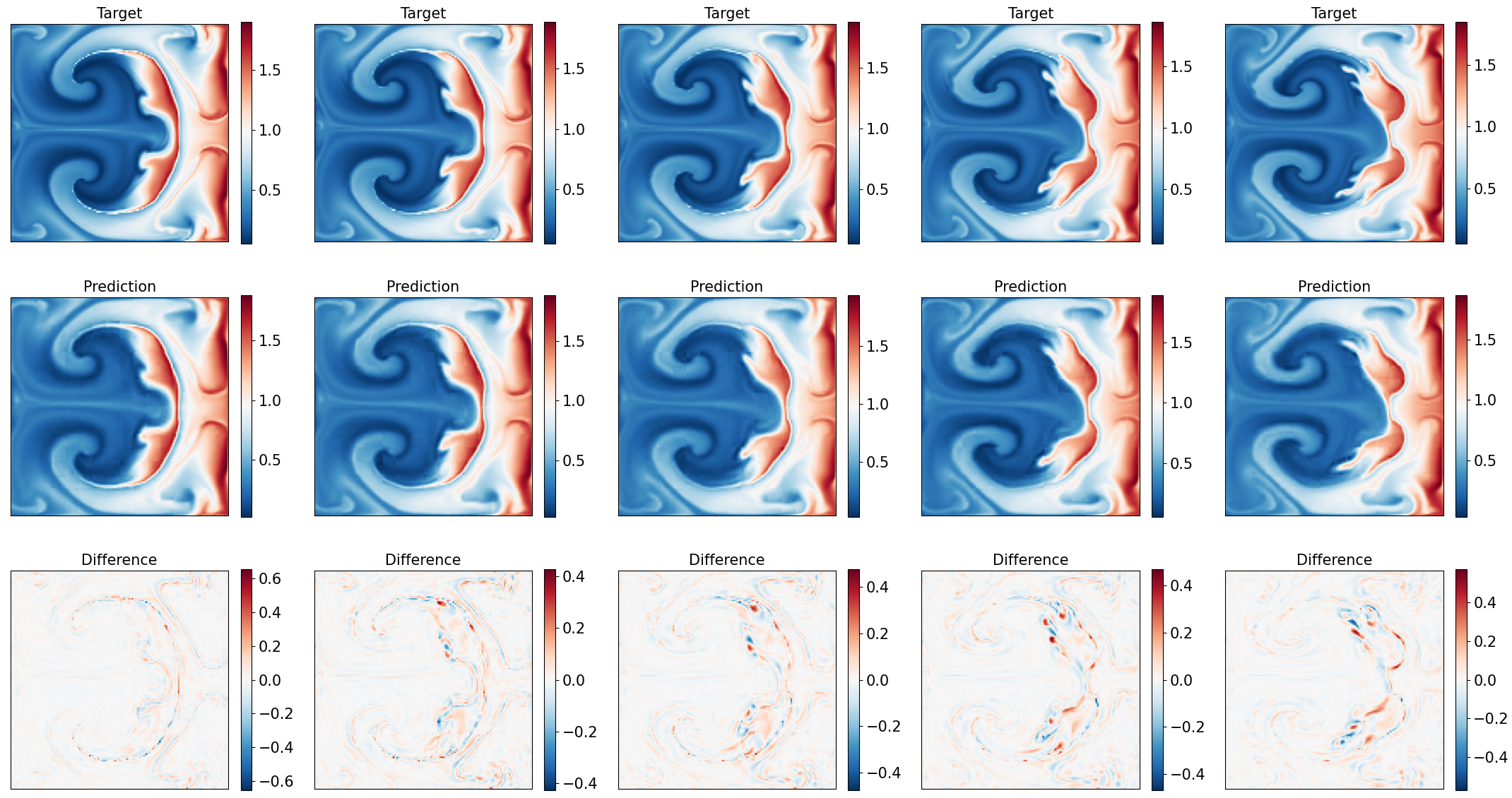}
   \caption{The channel plotted is the particle density.}
\end{subfigure}

\caption{\textbf{Example outputs for the PROSE-FD model.} 5 output steps for PDEArena NS-cond dataset (all three channels in equation \eqref{eq:arena_ns_c}). Each column represents a different timestep. For this trajectory, the relative $L^2$ error is 8.74\%.}
\label{fig:more_output}
\end{figure}

See Figure \ref{fig:more_output} for additional PROSE-FD model output visualizations. 

\end{document}